\documentclass[sigconf]{acmart}

\AtBeginDocument{%
  \providecommand\BibTeX{{%
    \normalfont B\kern-0.5em{\scshape i\kern-0.25em b}\kern-0.8em\TeX}}}

\usepackage{booktabs} 

\usepackage[ruled]{algorithm2e} 
\usepackage{graphicx}
\SetAlFnt{\small}
\SetAlCapFnt{\small}
\SetAlCapNameFnt{\small}
\SetAlCapHSkip{0pt}
\IncMargin{-\parindent}
\usepackage{mathrsfs}

\def\x{{solution}}
\def\xr{{recommended solution}}
\def\y{{output }}
\def\ys{{outputs }}

\def\bx{\bold{x}} 
\def\by{\bold{y}} 
\def\xs{{solutions}}

\def\SolSpace{{solution set }}

\def\ksteps{{p}}
\def\E{\mathbb{E}} 
\def\f{{objective function }}
\def\R{\mathbb{R}} 
\def\B{{B }}

\def\Fn{$\mathscr{F}^n$} 
\def\Un{$\mathscr{U}^n$} 
\def\f{{objective function }}
\def\btheta{\boldsymbol\theta}
\def\bf{\bold{f}}
\def\bY{\bold{Y}}
\DeclareMathOperator*{\amax}{arg\,max}
\def\Cov{\text{Cov}\,}
\begin{document}


\title{One Step Preference Elicitation\\ in Multi-Objective Bayesian Optimization}
%


\author{Juan Ungredda}
\email{J.Ungredda@warwick.ac.uk}
\affiliation{%
 \institution{University of Warwick}
 \streetaddress{}
 \city{Coventry}
 \state{}
 \country{UK}
 \postcode{CV4 7AL}
}

\author{Juergen Branke}
\email{juergen.branke@wbs.ac.uk}
\affiliation{%
 \institution{University of Warwick}
 \streetaddress{}
 \city{Coventry}
 \state{}
 \country{UK}
 \postcode{CV4 7AL}
}

\author{Mariapia Marchi}
\email{marchi@esteco.com}
\affiliation{%
 \institution{ESTECO SpA}
 \streetaddress{}
 \city{Trieste}
 \state{}
 \country{Italy}
 \postcode{}
}
\author{Teresa Montrone}
\email{montrone@esteco.com}
\affiliation{%
 \institution{ESTECO SpA}
 \streetaddress{}
 \city{Trieste}
 \state{}
 \country{Italy}
 \postcode{}
}

\renewcommand{\shortauthors}{}

\begin{abstract}
 
We consider a multi-objective optimization problem with objective functions that are expensive to evaluate. The decision maker (DM) has unknown preferences, and so the standard approach is to generate an approximation of the Pareto front and let the DM choose from the generated non-dominated designs. However, especially for expensive to evaluate problems where the number of designs that can be evaluated is very limited, the true best solution according to the DM's unknown preferences is unlikely to be among the small set of non-dominated solutions found, even if these solutions are truly Pareto optimal. We address this issue by using a multi-objective Bayesian optimization algorithm and allowing the DM to select a preferred solution from a predicted continuous Pareto front just once before the end of the algorithm rather than selecting a solution after the end. This allows the algorithm to understand the DM's preferences  and make a final attempt to identify a more preferred solution. We demonstrate the idea using ParEGO, and show empirically that the found solutions are significantly better in terms of true DM preferences than if the DM would simply pick a solution at the end.

 
\end{abstract}



    \maketitle

\section{Introduction}
Many real-world optimization problems have multiple, conflicting objectives. A popular way to tackle such problems is to search for a set of Pareto-optimal solutions with different trade-offs, and allow the decision maker (DM) to pick their most preferred solution from this set. This has the advantage that the DM doesn't have to specify their preferences explicitly before the optimization, which is generally considered very difficult.

In this paper, we consider the case of expensive multi-objective optimization problems, where the evaluation of the objective functions is very costly, and thus the number of solutions that can be evaluated during optimization is small. This creates challenges for optimization, but also means that the Pareto front, which may consist of thousands of Pareto-optimal solutions or even be continuous, can only be approximated by a small set of solutions. It is thus unlikely that the solution most preferred by the DM would be among the small set of solutions found by the optimization algorithm, even if these are among the true Pareto-optimal solutions.

We suggest to tackle this issue by using Bayesian Optimization (BO), a surrogate-based global optimization technique. BO is not only known to be very suitable to expensive optimization as it carefully selects points to evaluate through an acquisition function that explicitly balances exploration and exploitation. It also generates a surrogate model of each objective function. These surrogate models can be optimized by a multi-objective evolutionary algorithm to generate an approximated Pareto front, and as evaluation of the surrogate model is cheap relative to a fitness evaluation, we can generate a fine-granular representation of the approximated Pareto front, consisting of very many solution candidates.
This approximated Pareto front with many hypothetical solutions can then be shown to a DM to select from. While we cannot guarantee that the picked solution is actually achievable, the location of the picked solution should still give us a very good idea about the DM's preferences. Essentially, it provides a reference point which we suspect to be very close to the true Pareto front. We then continue to run BO for a few more steps (often only one more step), aiming to generate the desired solution or something better. We demonstrate this idea by using ParEGO \cite{Know06a}, and show empirically that this preference information allows to generate a significantly better final solution.

We believe that the cognitive burden for the DM is not much higher than in standard multi-objective optimization: rather than having to identify the most preferred solution from a discrete approximation of the Pareto front at the end of the run, they now pick the most preferred out of the predicted (larger) set of solutions, but the size of the set presented to the DM should not make a big difference in terms of cognitive effort if the problem has only 2 (perhaps 3) objectives, where the interesting region can be identified easily by inspecting the Pareto front visually. After the final optimization step, if we have found a solution that dominates the DM's chosen solution, there is no need for the DM to choose again - we know that the new found solution is the most preferred one. If no such new solution is found in the final step, the DM would have to look at the set of found solutions again (this time the small set of actually evaluated non-dominated solutions) and pick a final solution. However, given that the DM has already engaged with the approximated Pareto front, also this selection step should carry a low cognitive load.

We examine the trade-off with respect to when the DM is asked to pick a solution. If the DM is asked too early, the approximation of the Pareto front generated may be poor, and the preference information learned not very reliable. On the other hand, if the DM is asked too late, the algorithm has not many attempts to find a better solution. We find that while there is already a significant benefit in letting the DM choose just one iteration before the end of the algorithm. Asking earlier may yield even greater benefits in case the predicted Pareto front is already reasonable accurate, but may be detrimental if the budget of evaluations is very small and no good approximation has been found yet.

Our paper is structured as follows. Section~\ref{sec:relatedWork} briefly summarizes some related work. The problem we consider is formally defined in Section~\ref{sec:problemDefinition}. Section~\ref{sec:algorithm} describes the proposed algorithm, and empirical results on several benchmark problems are presented in Section~\ref{sec:results}. The paper concludes with a summary and some ideas for future work.

\section{Literature Review\label{sec:relatedWork}}

Depending on the involvement of the DM in the optimization process, multi-objective optimization can  be classified into a priori approaches, a posteriori approaches, and interactive approaches \citep{branke2016mcda,Xin18}. The field is very large, so we can only mention some of the most relevant papers here. A priori approaches ask the DM to specify their preferences ahead of optimization. This allows to turn the multi-objective optimization problem into a single objective optimization problem, but it is usually very difficult for a DM to specify their preferences before having seen the alternatives. One effective way to capture preferences is to ask the DM to specify a reference point \citep{DEB2006}. 

Most multi-objective EAs are a posteriori approaches, attempting to identify a good approximation of the Pareto frontier, and the DM can then pick the most preferred solution from this set. This is much easier for a DM, but identifying the entire Pareto front may be computationally expensive.

Interactive approaches attempt to learn the DM's preferences during optimization and then focus the search on the most preferred region of the Pareto front. This often involves asking the DM to compare pairs of solutions, and usually assumes some utility function model.  The achievement scalarizing function based on the Tchebychev utility is widely used. It has the advantage that also solutions in concave regions of the Pareto front are supported \citep{Wierzb77}. 
However, also more flexible models based on robust ordinal regression have been used \citep{BGSZ09}.

Our proposed algorithm lies in between a priori and interactive approaches: It generates an approximation of the Pareto front a priori, and only requires the DM to pick a solution from this front, but it may then make a final attempt to find a more preferred solution based on what the DM has picked.

Bayesian optimization is a global optimization technique that builds a Gaussian process (GP), often called Kriging, surrogate model of the fitness landscape, and then uses the estimates about mean and variance at each location to decide which solution to evaluate in the next iteration. It uses an acquisition function to explicitly make a trade-off between exploitation and exploration (see, e.g., \cite{NandoDeFreitas}). A frequently used acquisition function is the expected improvement (EI) \cite{jones1998efficient} which selects the point with the largest expected improvement over the current best known solution as the next solution to evaluate. Recently, Bayesian optimization has been adapted to the multi-objective case, for a survey see \cite{ROJASGONZALEZ2020}. One of the earliest approaches, ParEGO \citep{Know06a} simply uses the Tchebychev achievement scalarization function to turn the multi-objective problem into a single objective problem, but uses a different scalarization vector in every iteration where the next \x\ is collected according to EI. Other multi-objective algorithms fit separate models for each individual objective. \cite{Emmerich2011} trains a Gaussian process model for each objective function, then chooses the next solution to be evaluated according to a hypervolume-based acquisition criterion. Other multi-objective BO approaches include \cite{keane06,picheny15}.

An interactive multi-objective BO approach has recently been proposed in \cite{AstudilloFrazier20}. The approach assumes the DM has a linear utility function and regularly asks the DM to rank pairs of solutions  to narrow down the search onto the most interesting region. 

We are aware of only one paper that attempts to use reference points in multi-objective Bayesian optimization. \cite{gaudrie2020} modify the expected hypervolume improvement criterion to take into account a reference point and focus the search on the preferred region. This could be an alternative approach to the reference direction approach we use in this paper for the final phase of the optimization. The DM can also propose reference points in \cite{Pour21}, and a Gaussian process is used in combination with multi-objective evolutionary algorithms for optimization. The acquisition function balances DM preferences and model uncertainty.

\section{Problem Definition\label{sec:problemDefinition}}

The standard multi-objective optimization problem with respect to a particular DM is defined as follows.
We assume a $D$-dimensional  real-valued space of possible \textit{\xs}, i.e., $\bx \in X \subset \R^{D}$. The \f is an arbitrary black box $\bf: X \to \R^{K}$ which takes as arguments a \x\ and returns a deterministic vector \textit{\y} $\by \in \R^{K}$. The (unknown) DM preference over the \ys can be characterized by a utility function $U:\R^{K} \to \R$. Thus, of all \xs\ in $X$, the DM's most preferred solution is $\bx^{*} = \amax_{\bx\in X} U(\bf(x))$. 


There is a budget of $\B$ objective function evaluations, and we denote the $n$-th evaluated design by $\bx^{n}$ and the $n$-th \y by $\by^{n} = \bf(\bx^n)$ where, for convenience, we define the sampled \xs\ and \ys as $\tilde{X}^n = \{\bx^{1},\dots,\bx^{n}\}$ and $\tilde{\bY}^{n} = \{\by^1,\dots, \by^n\}$, respectively. 

\begin{figure}
	\centering
	\begin{tabular}{c}
		\includegraphics[height=8.0cm]{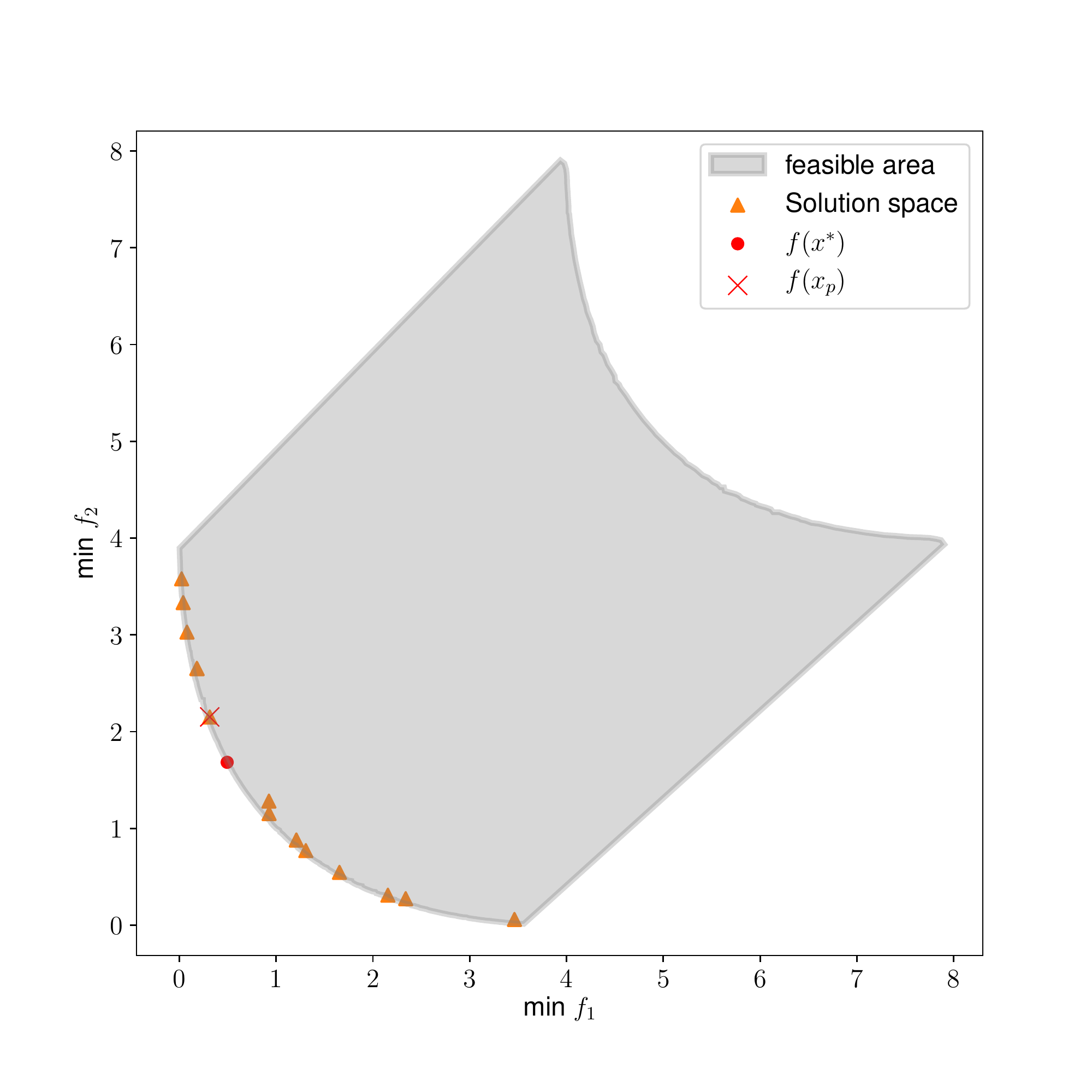}\\
	\end{tabular}
    	\captionof{figure}{HOLE test problem (b=0). (orange triangles ) Solution space $\Gamma$ shown to the DM. (red cross) Solution picked by the DM $x_{p}$. (red dot) Solution that the DM would have picked in the front. }
	\label{fig:POL_Problem_Description}
\end{figure}

After consuming the budget, the DM chooses a preferred solution $\bx_{p}$ from a \SolSpace $\Gamma \subset \tilde{X}^{n}$ according to  $\bx_{p}=\amax_{x \in \Gamma} U(\bf(x))$. Figure \ref{fig:POL_Problem_Description} shows that a \SolSpace without information from the DM may result in clear differences between the preferred solution, $\bx^{*}$, and the preferred sampled solution, $\bx_{p}$. The quality of the solution set $\Gamma$  is determined by the utility of the chosen sample, and we choose to minimize the Opportunity Cost (OC) or regret:
$$
OC = U(\bf(\bx^{*})) - U(\bf(\bx_{p}))
$$
We assume that, rather than just presenting the DM the \SolSpace $\Gamma$ after the optimization, it is possible to engage with the DM once after $\B-\ksteps$ evaluations. This information can then be used to focus the search in the final $\ksteps$ evaluations, hopefully allowing to generate a \SolSpace $\Gamma$ better tailored to the DM's preferences and thus having a lower $OC$.  

\section{Algorithm\label{sec:algorithm}}
While we believe that the basic idea is compatible with various BO algorithms, we consider ParEGO \cite{Know06a} as a multi-objective optimization algorithm here (see Section~\ref{ParEGO}). At step $m=B-\ksteps$, the DM is presented with an (effectively) continuous approximation of the Pareto front. The solution chosen yields information about the direction of optimization in the final $\ksteps$ optimization steps. In Subsection \ref{sec:StatisticalModelUtility}  we describe the Gaussian process model. Subsection \ref{ParEGO} describes briefly ParEGO. Lastly, Subsection \ref{ParEGO with Last Preferential Step} explains details of the proposed approach.

\subsection{Statistical Model over Simulator and Utility}\label{sec:StatisticalModelUtility}

Let us denote the set of evaluated points and their \f up to iteration $n$ as \Fn $= \{(\bx,\by)^1,\dots,(\bx,\by)^n \}$, and the utility of each \x\ as \Un $= \{(\bx,u)^1,\dots,(\bx,u)^n \}$, with $u=U(\bf(\bx))$.
 Specifically, we define the sampled \y data for the $j$-th \f as $\tilde{Y}_{j}^{n}$. Then, we propose to use an independent GP to model each objective function $y_{j}=f_{j}(\bx) ,\forall j=1,\dots,K$, defined by a mean function $\mu_{j}^0(\bx):X \to \R$ and a covariance function $k_{j}^0(\bx, \bx'):X \times X \to \R$. Given the  dataset \Fn, predictions at new locations $\bx$ for \y $y_{j}$ are given by 
\begin{align*}
\begin{split}
\mathbb{E}[f_{j}(\bx)] &= \mu_{j}^n(\bx) \\
&=\mu_{j}^0(\bx)-k_{j}^0(\bx,\tilde{X}^n)k_{j}^0(\tilde{X}^n,\tilde{X}^n)^{-1}(Y_{j}^n-\mu_{j}^0(\tilde{X}^n))
\end{split}\\
\begin{split}
\Cov[f_{j}(\bx),f_{j}(\bx')] &= k_{j}^n(\bx,\bx')\\
&= k_{j}^0(\bx,\bx')-k_{j}^0(\bx,\tilde{X}^n)k_{j}^0(\tilde{X}^n,\tilde{X}^n)^{-1}k_{j}^0(\tilde{X}^n,\bx')
\end{split}
\end{align*}


The prior mean $\mu_{j}^0(\bx)$ is typically set to $\mu_{j}^0(\bx)=0$ and the kernel $k_{j}^0(\bx, \bx')$
allows the user to encode known properties of the \f $f_{j}(\bx)$ such as smoothness
and periodicity. We use the popular squared exponential kernel that assumes
$f_{j}(\bx)$ is a smooth function such that nearby $\bx$  have similar outputs while widely separated points have unrelated outputs,
\begin{align*}\label{kernel}
k_{j}^{0}(\bx,\bx') = \sigma_{0}^{2}\exp\left( \frac{||(\bx)-(\bx')||^{2}}{2l_{X}^{2}}\right) 
\end{align*}
where $\sigma_{0} \geq 0$ and $l_{X} > 0$ are hyper-parameters estimated from the data \Fn\ by maximum marginal likelihood. Further details can be found in \cite{Rasmussen06}.

Let us assume that the DM's utility can be described with a parametric utility $U_{\btheta}(\bx)$ with parameters $\btheta \in \Theta$. Then, similarly to modeling an objective function $\bf$, we may use a GP to model $U_{\btheta}$ by a mean function $\mu_{U}^0(\bx):X \to \R$ and a covariance function $k_{U}^0(\bx, \bx'):X \times X \to \R$. 

\subsection{ParEGO\label{ParEGO}}

  ParEGO \citep{Know06a} translates a multi-objective problem into a single-objective problem using the following achievement scalarization function based on the Tchebychev function. 
 
 $$
U_{\btheta}(\bx) = \max_{j=1,\dots,K}(\theta_{j}f_{j}(\bx)) + \rho \sum_{j=1,\dots,K} \theta_{j}f_{j}(\bx)
$$
 
 where $\rho$ is usually a very small number.  The parameter $\btheta \in \Theta$ is defined as, 

$$
\btheta = \{\btheta \in [0,1]^{K}| \sum_{j}\theta_{j}=1 \}
$$
 

An iteration of ParEGO consists of randomly sampling a parameter vector $\btheta$ and compute the dataset \Un\ based on \Fn using $U_{\btheta}(\bx)$. Then, \Un\ is used to build a GP metamodel where the next \xr, $x_{r}$, is given by the \x\ that maximizes the expected improvement,

$$
EI_{\btheta}(x) = \E[\max{\big(U_{\btheta}(x) -u_{max},0\big)}]
$$
where,
$$
u_{max} = \max_{i=1,\dots, n} U_{\btheta}(x^{i})
$$

This process is repeated until the budget \B has been depleted.

\subsection{ParEGO with One Step Preference Elicitation}\label{ParEGO with Last Preferential Step}

ParEGO  returns a set of non-dominated solutions. But especially for expensive to evaluate problems where the number of evaluated solutions is extremely limited, the true most preferred solution $x^{*}$ may be quite far away from any found non-dominated solution, even if these are truly Pareto-optimal. This is because even a set of Pareto optimal solutions will have gaps that can contain the true most preferred solution (cf.~Figure~\ref{fig:POL_Problem_Description}).
To focus more directly on the region interesting to the DM, at step $m=B-\ksteps$, we fit a GP for each \f $f_{j}(x)$, using $\mathscr{F}^m$ to produce a response surface over each \y using the mean posterior $\mu^{m}_{j}$. Then, we use NSGA-II to produce a Pareto front approximation based on the response surfaces and show this approximation to the DM. 

Assuming that the DM's utility function is based on the Tchebychev utility, we can estimate the parameters $\btheta$ from the solution the DM has picked, $x_p$, by

$$
\frac{\hat \theta_i}{\hat \theta_j}=\frac{f_j(x_p)}{f_i(x_p)}\quad\quad \forall i,j=1\ldots K
$$





\begin{figure}
	\centering
	\begin{tabular}{c}
		\includegraphics[height=8.0cm]{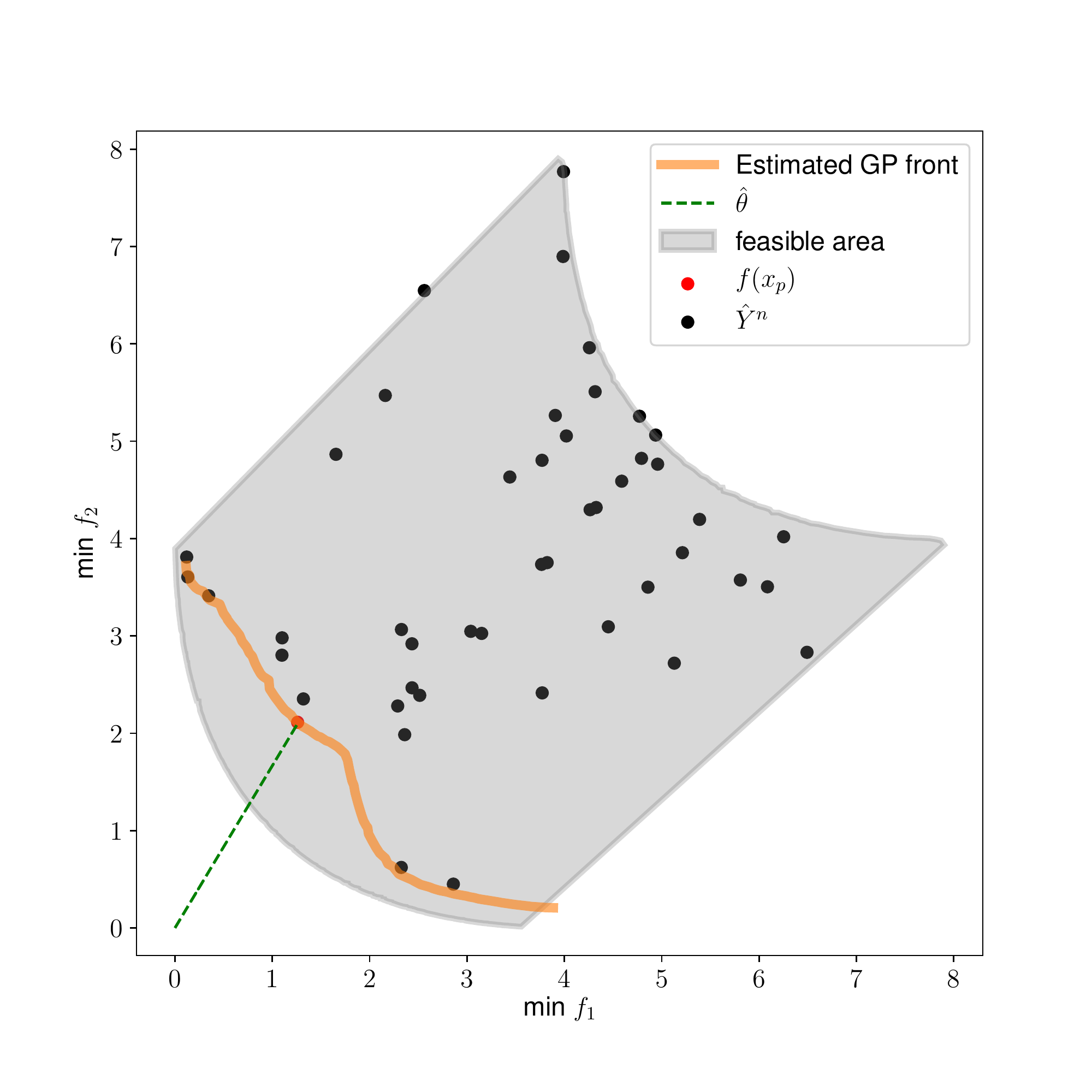}\\
	\end{tabular}
    	\captionof{figure}{HOLE test problem (b=0). (orange) Estimated Pareto front based on the mean posterior $\mu^{n}_{j}$. (black) sampled output data. (red) Solution chosen by the DM in the front. (green) estimated direction given $\hat{\theta}$.  }
	\label{fig:POL_Alg_Description}
\end{figure}

Figure \ref{fig:POL_Alg_Description} shows the proposed approach at step $m$. Finally, we compute the last $\ksteps$ optimization steps by using expected improvement with the estimated parameters $EI_{\hat{\btheta}}(x)$. 

\section{Results and Discussion\label{sec:results}}

To demonstrate the performance of the proposed approach, we compare it against using ParEGO for the full budget of $B$ evaluations, without exploiting information from a DM in the final $\ksteps$ optimization steps.  

To plot the convergence of the opportunity cost (OC) over iterations, we want to show, at each iteration $i$, the performance if the algorithm would have stopped there.
Thus, to determine the OC at iteration $i$, we take the set of non-dominated solutions generated up to iteration $i-\ksteps$, create an approximation of the Pareto front, select the most preferred solution from that approximated front by using the true DM's preferences, and finally execute iterations $i-\ksteps+1,\ldots, i$ of the optimization taking the preference information into account. The final OC at iteration $i$ is then the OC of this final solution set and is based on the true DM utility function.


\subsection{Experimental setup}

We assume the DM has a Tchebychev utility with true underlying (but unknown to the algorithm) parameters $\theta$. The true underlying parameters are generated randomly for every replication of a run using a different random seed. 
In all experiments, NSGA-II is run for 300 generations with a population size of $100$ to produce a Pareto front approximation.

We use three different test functions: The HOLE function \cite{Rigoni2004} is defined over $X = [-1,1]^{2}$ and has $K=2$ objectives. We show results for two parameter settings $b > 0$ (denoted HOLE-1) and $b=0$ (denoted HOLE-2). In the first setting, a hole is produced which creates two unconnected Pareto fronts whereas for the latter a continuous convex Pareto front is generated. The POL function \cite{Poloni1997} is defined over $X = [-\pi,\pi]^{2}$ and also has $K=2$ objectives. Details for all three functions can be found in the appendix. All results are averaged over multiple independent replications (50 replications for HOLE functions and 160 replications for POL).

\subsection{Benefit of Final Step Preference Elicitation}
In this subsection, we look at the benefit that can be gained from showing the DM the approximated Pareto front at the final stage of optimization, with $\ksteps=1$ or $\ksteps=2$.

\begin{figure}
	\centering
		\begin{tabular}{c}
		\includegraphics[height=3.8cm]{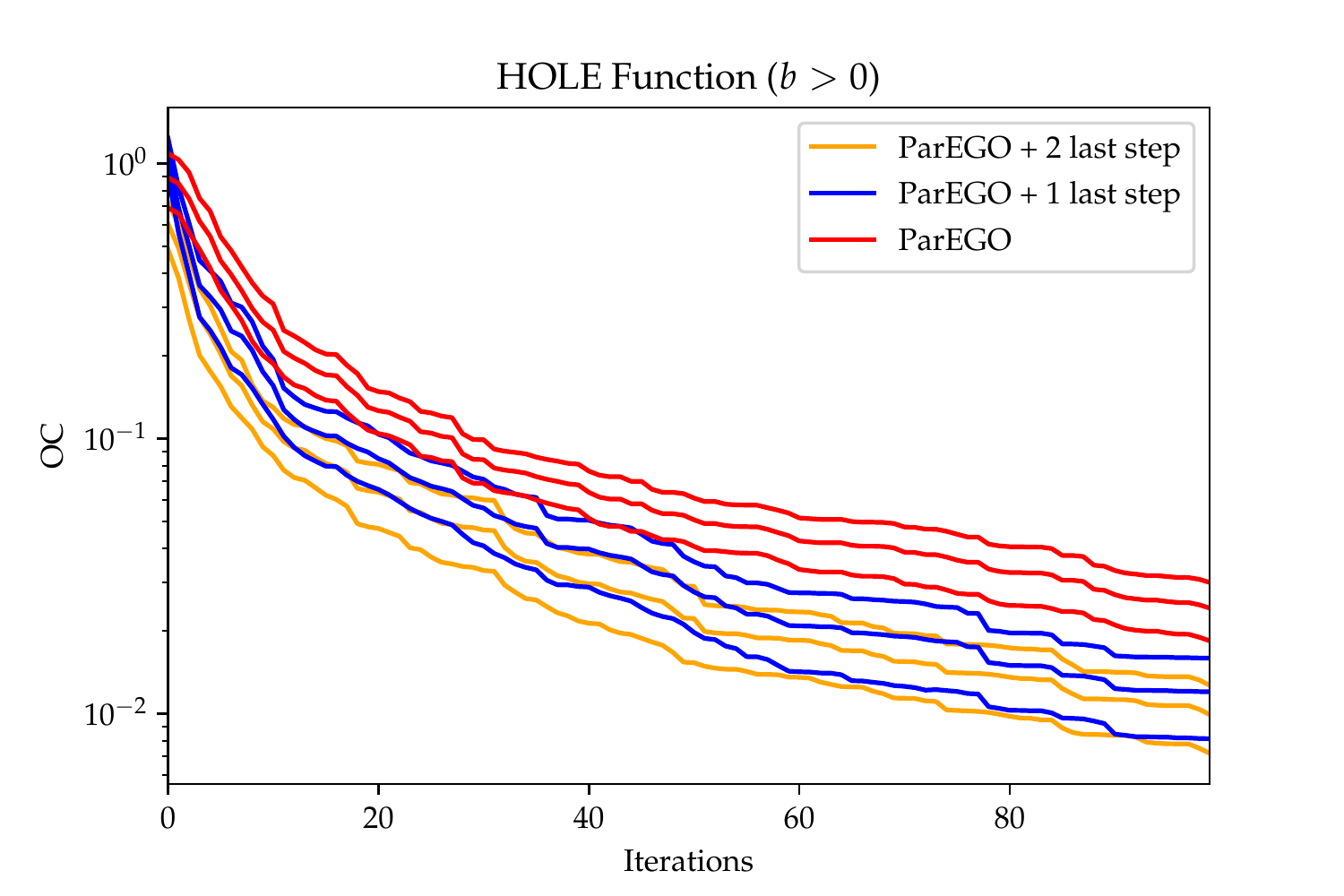}\\
		(a)\\
			\includegraphics[height=3.8cm]{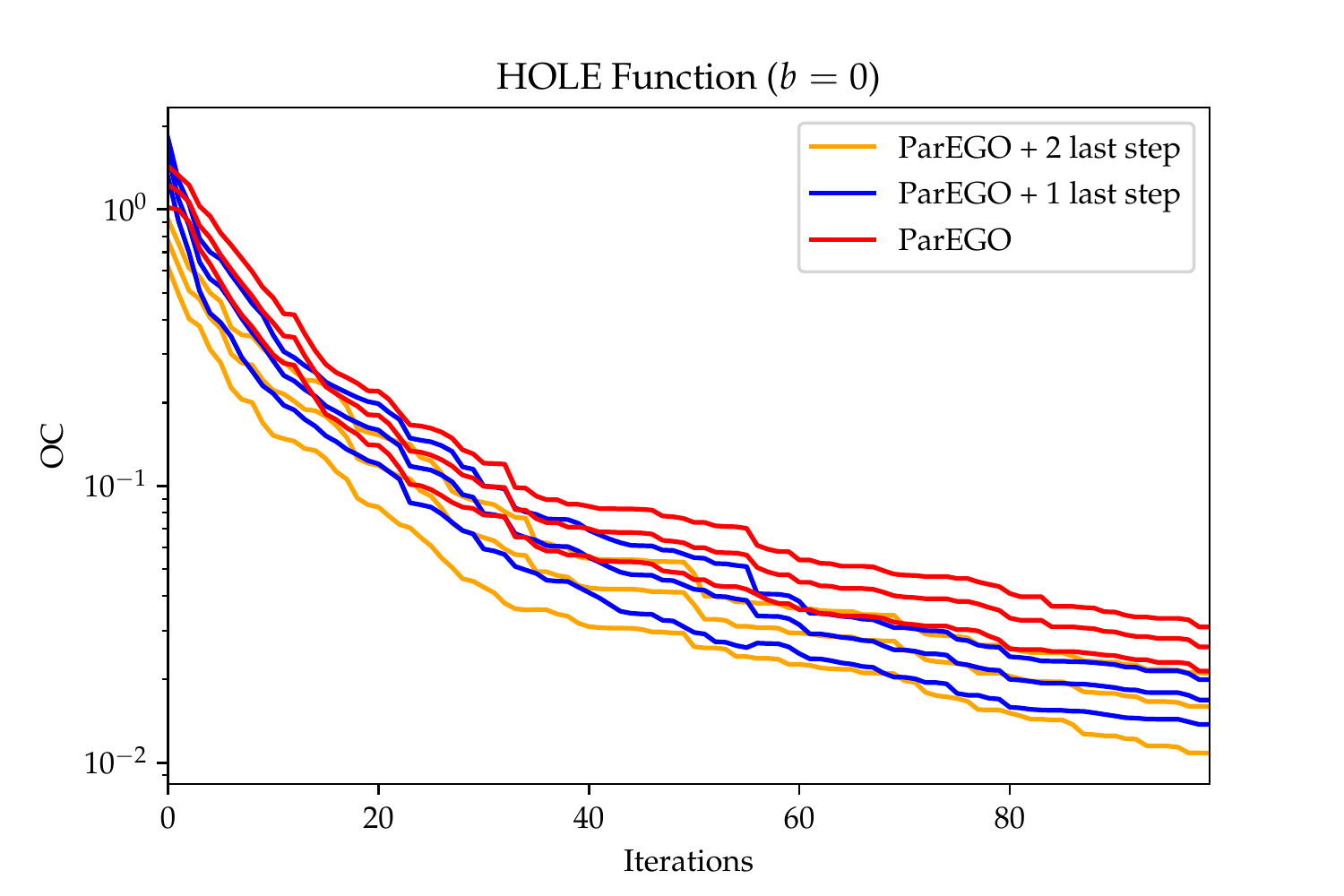}\\
		(b)\\
		\includegraphics[height=3.8cm]{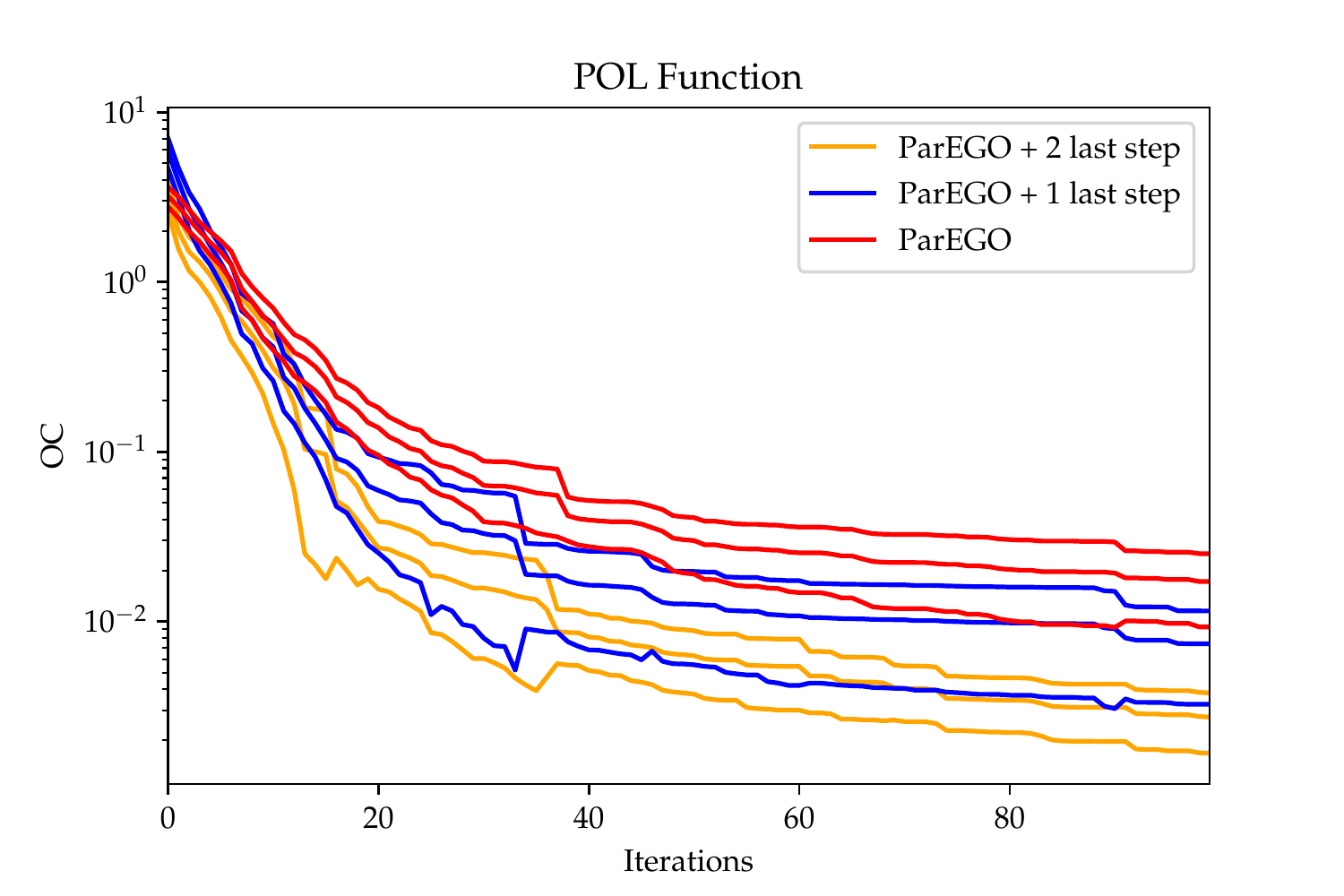}\\
		(c)
		    	\end{tabular}
    	\captionof{figure}{Comparison of ParEGO and ParEGO with preference elicitation for $p=1$ and $p=2$. Mean and 95\% CI for the OC over iterations. (a) HOLE-1 function with $b>0$, (b) HOLE-2 function with $b=0$, (c) POL function.}
	\label{fig:HOLE_RESULTS}
\end{figure}

Figures~\ref{fig:HOLE_RESULTS} (a) and (b) show, for both HOLE test functions, that the proposed approach allows to find significantly better solutions (in terms of true DM utility) than if the DM is only given the final solution set to choose from. Whether the DM is consulted only one $(\ksteps=1)$ or two steps ($\ksteps=2$) before the end of the run does not seem to make a significant difference.

For the POL function (see Figure~\ref{fig:HOLE_RESULTS}(c)), results look similar, although there seems to be a more noticeable difference between the settings of $\ksteps=1$ and $\ksteps=2$.

\subsection{Mismatch with the DM's Utility model}
Our approach assumes the DM's utility to be a Tchebychev utility model, as this is the scalarization function used as default in ParEGO. 
In this section, we examine what happens if the DM's true utility model is different, meaning there is a mismatch between the DM's true utility model (unknown to the algorithm) and the learned Tchebychev model used in the final steps of ParEGO. 

So, we assume that the true DM utility model is \emph{linear}, and this linear model is used when the DM picks a solution from the approximated Pareto front at step $B-\ksteps$, and again to evaluate the OC at the end. ParEGO still derives a Tchebychev function and uses this in the final $\ksteps$ steps. As before, in each replication of the algorithm, a random  true utility model is generated for the DM, only that it is linear rather than Tchebychev.

\begin{figure}
	\centering
	\begin{tabular}{c}
		\includegraphics[height=3.8cm]{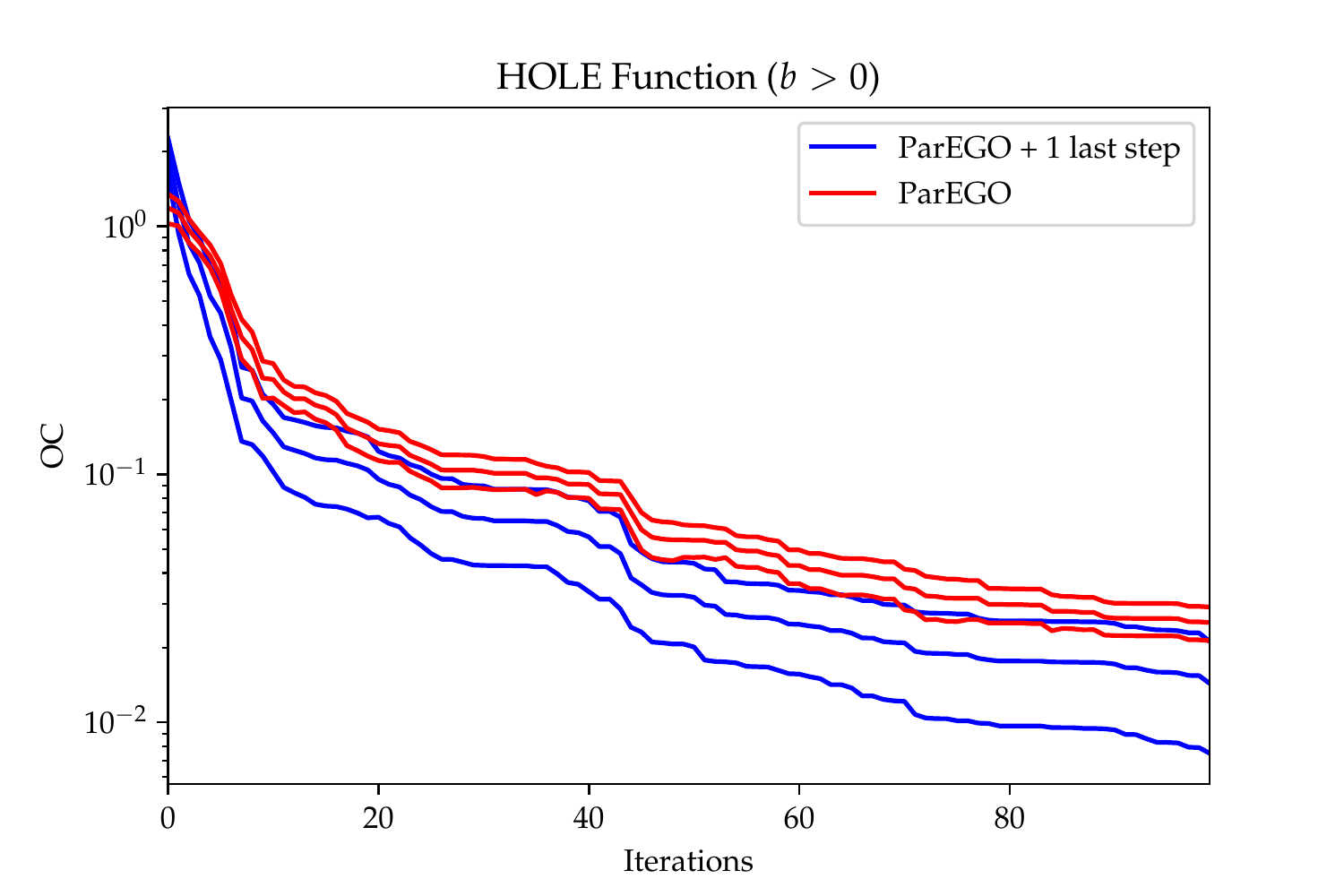}\\
		(a)\\
		\includegraphics[height=3.8cm]{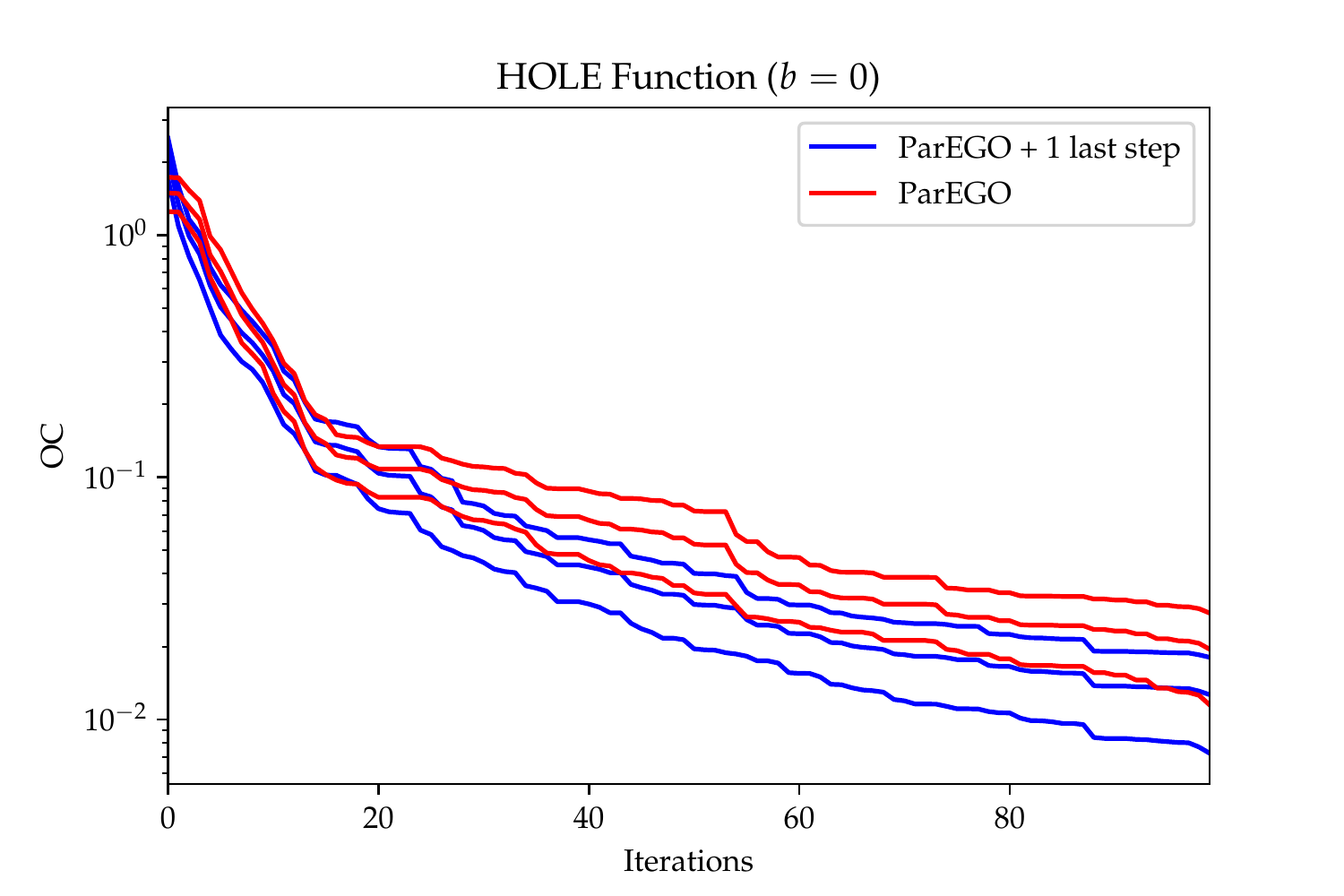}\\
		(b)\\
		\includegraphics[height=3.8cm]{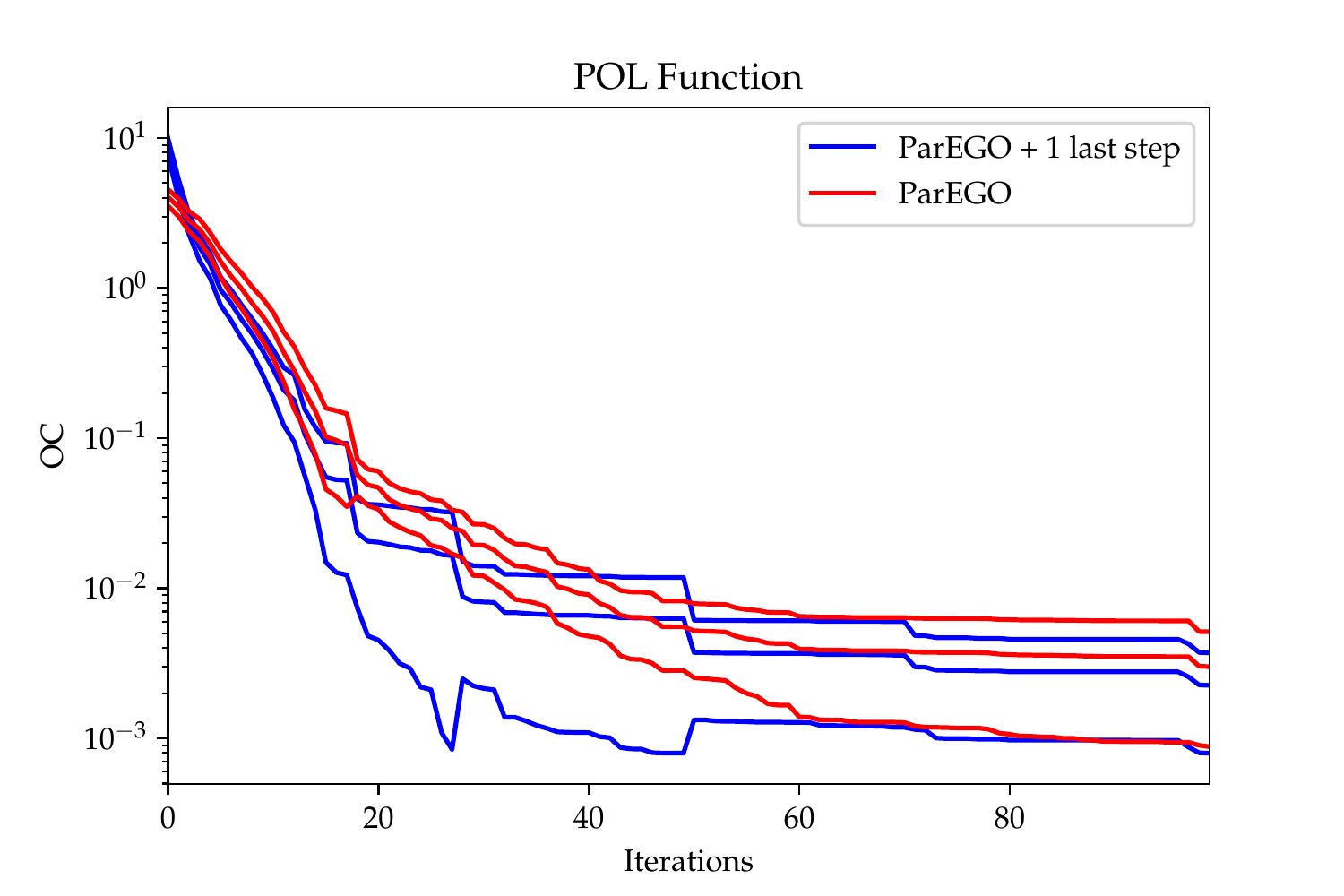}\\
		(c)
	\end{tabular}
    	\captionof{figure}{Comparison of ParEGO and ParEGO with preference elicitation for $p=1$. Mean and 95\% CI for the OC over iterations. True utility function of DM is \emph{linear}. (a) HOLE-1 function with $b>0$, (b) HOLE-2 function with $b=0$, (c) POL function.}
	\label{fig:HOLE_Model_Missmatch_RESULTS}
\end{figure}

As Figure~\ref{fig:HOLE_Model_Missmatch_RESULTS} shows, there is still a benefit from asking the DM one step before the end of the run, although it seems slightly smaller than without the utility model mismatch.

\subsection{Asking the DM earlier}
The larger benefit of $\ksteps=2$ over $\ksteps=1$ prompted us to consider even larger values of $\ksteps$. The earlier the DM is involved, the longer it is possible for ParEGO to exploit the preference information gained. On the other hand, the earlier the DM is shown an approximated Pareto front, the less accurate is this Pareto front, and thus the learned preference information may be wrong. Also, the difference between the approximated Pareto front shown to the DM early on and the final set of solutions identified will be larger, which means the additional cognitive effort to the DM is likely to be higher.

\begin{figure}[H]
	\centering
	\begin{tabular}{c}
		\includegraphics[height=3.8cm]{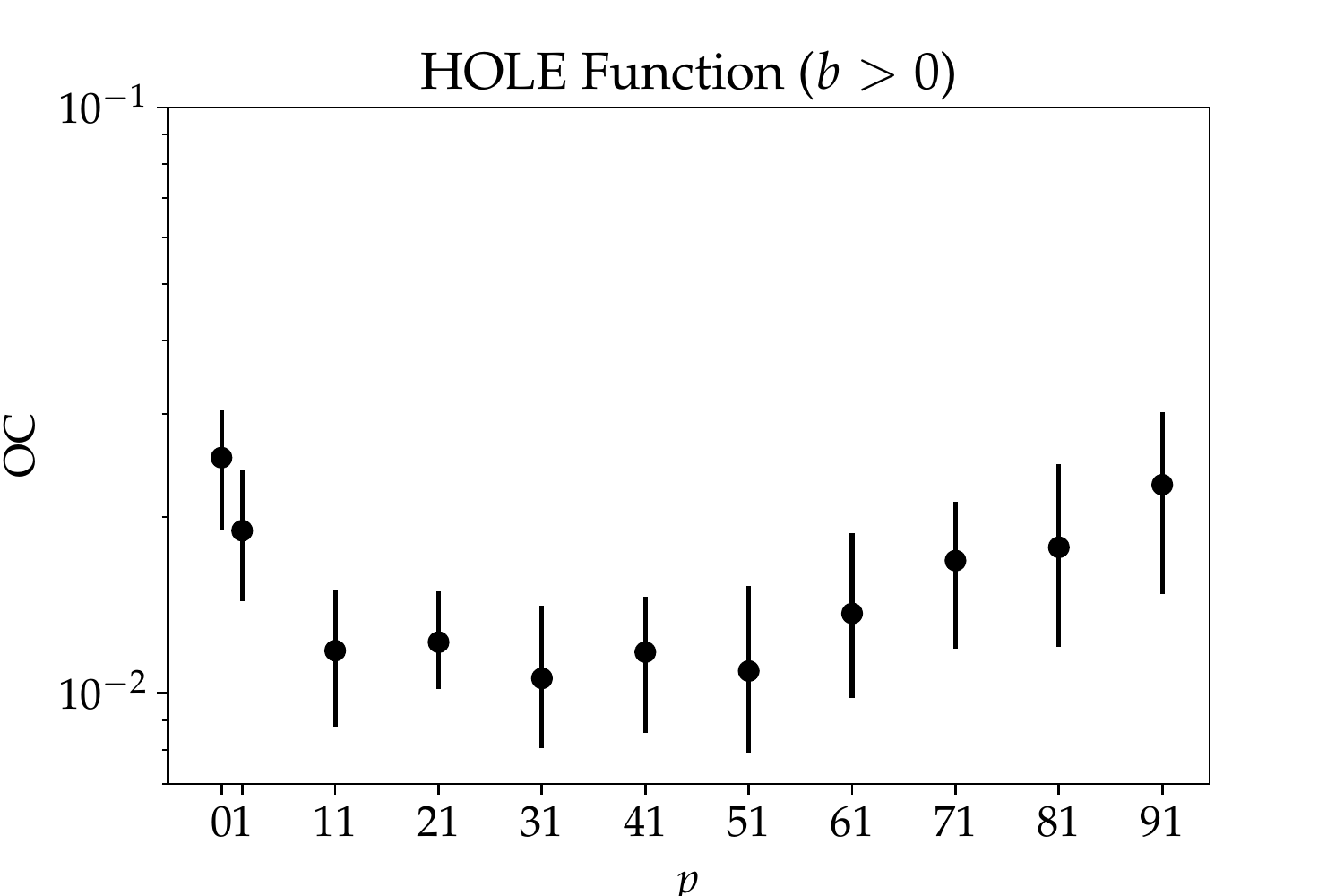}\\
		(a)\\
		\includegraphics[height=3.8cm]{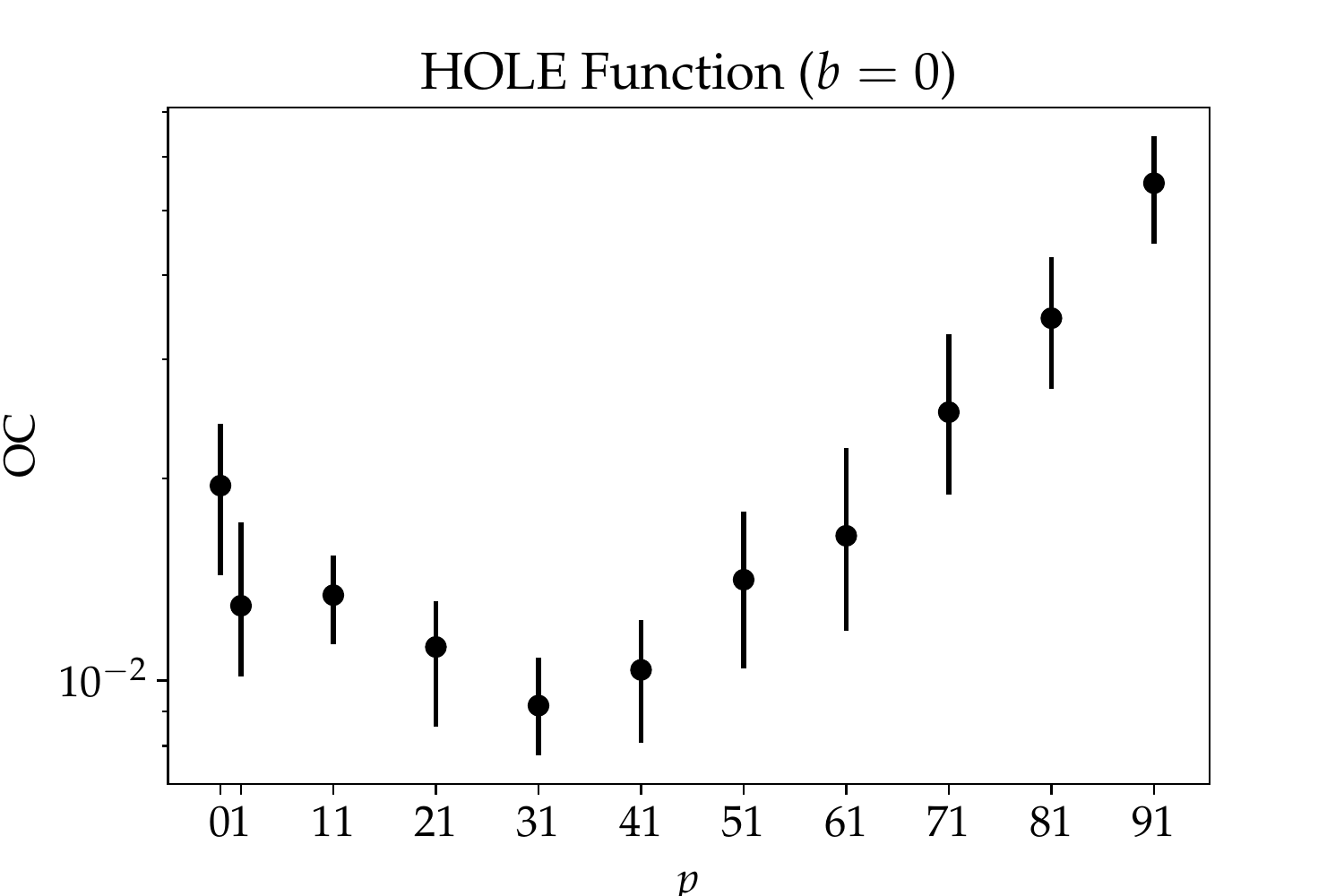}\\
		(b)\\
		\includegraphics[height=3.8cm]{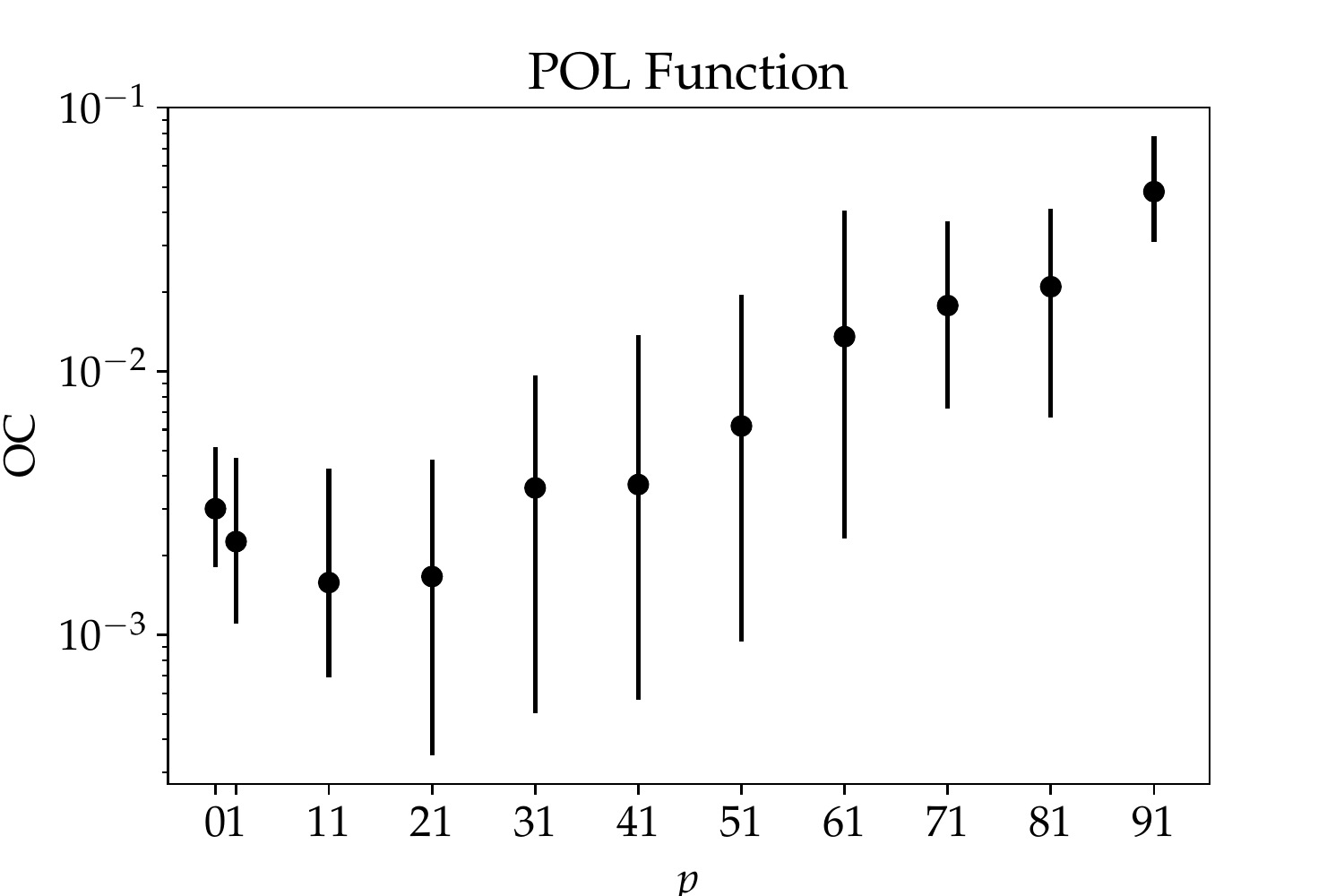}\\
		(c)\\
	\end{tabular}
    	\captionof{figure}{Final true utility of the generated solution set after $B=100$ iterations, depending on how many iterations before the end the DM was presented with an approximation of the front ($p$). True DM utility was \emph{linear}. Mean and 95\% CI for the OC.  (a) HOLE-1 function with $b>0$ (b) HOLE-2 function with $b=0$, (c) POL function.}
	\label{fig:HOLE_Ushape_PLot}
\end{figure}

Figure~\ref{fig:HOLE_Ushape_PLot} depicts the final utility to the DM depending on $\ksteps$, with $B=100$. As can be seen, for our experimental setting but over all three test functions, best results are obtained when the DM is shown the approximated front approximately 20 iterations before the end of the run (or after 80\% of the total budget). Therefore, after 80 evaluations, the approximated Pareto front seems of sufficient quality to make the estimated parameter $\hat{\btheta}$ close enough to $\btheta$ and guide the remaining optimization steps in a reasonable direction.

\begin{figure}[H]
	\centering
	\begin{tabular}{c}
		\includegraphics[height=3.8cm]{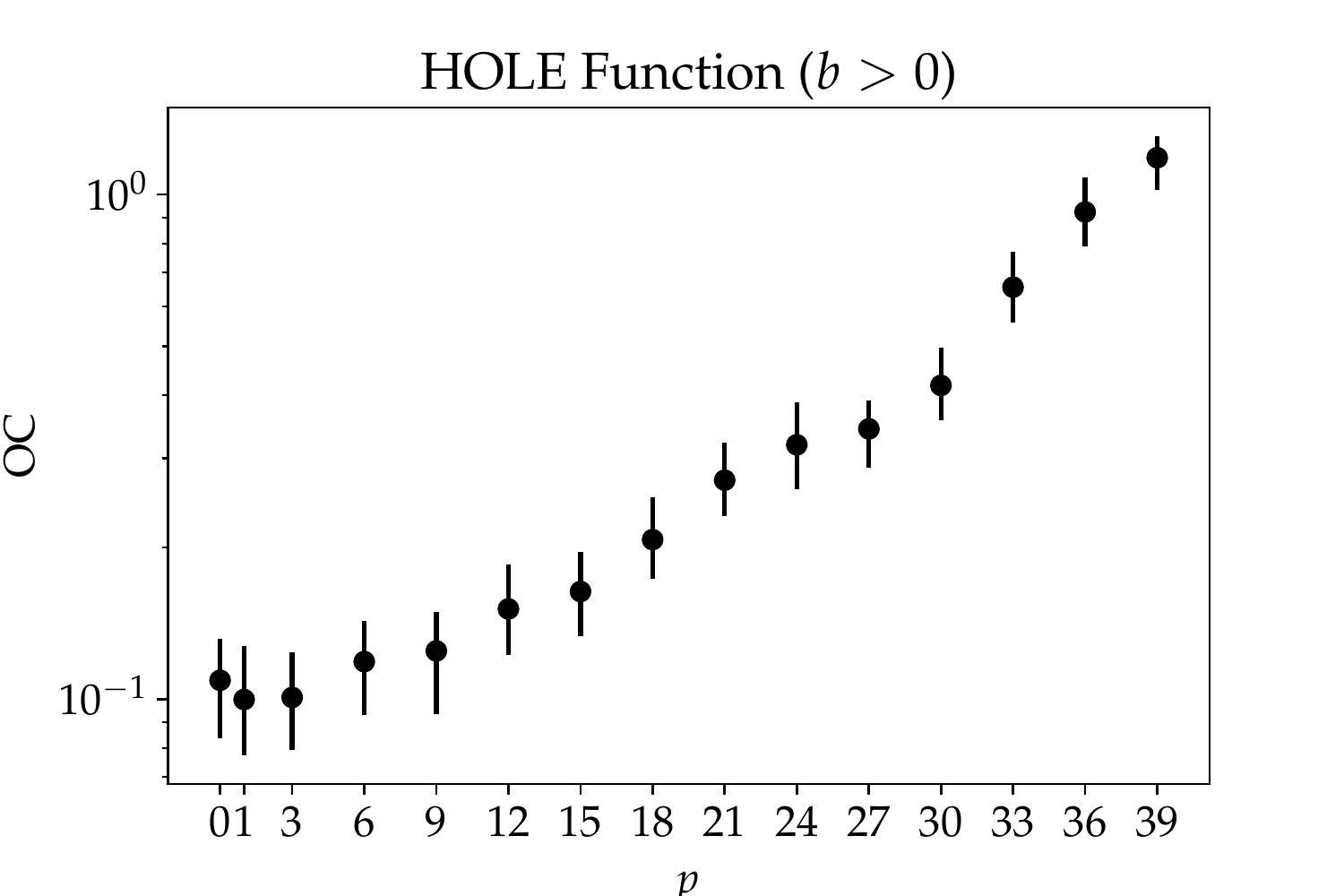}\\
		(a)\\
		\includegraphics[height=3.8cm]{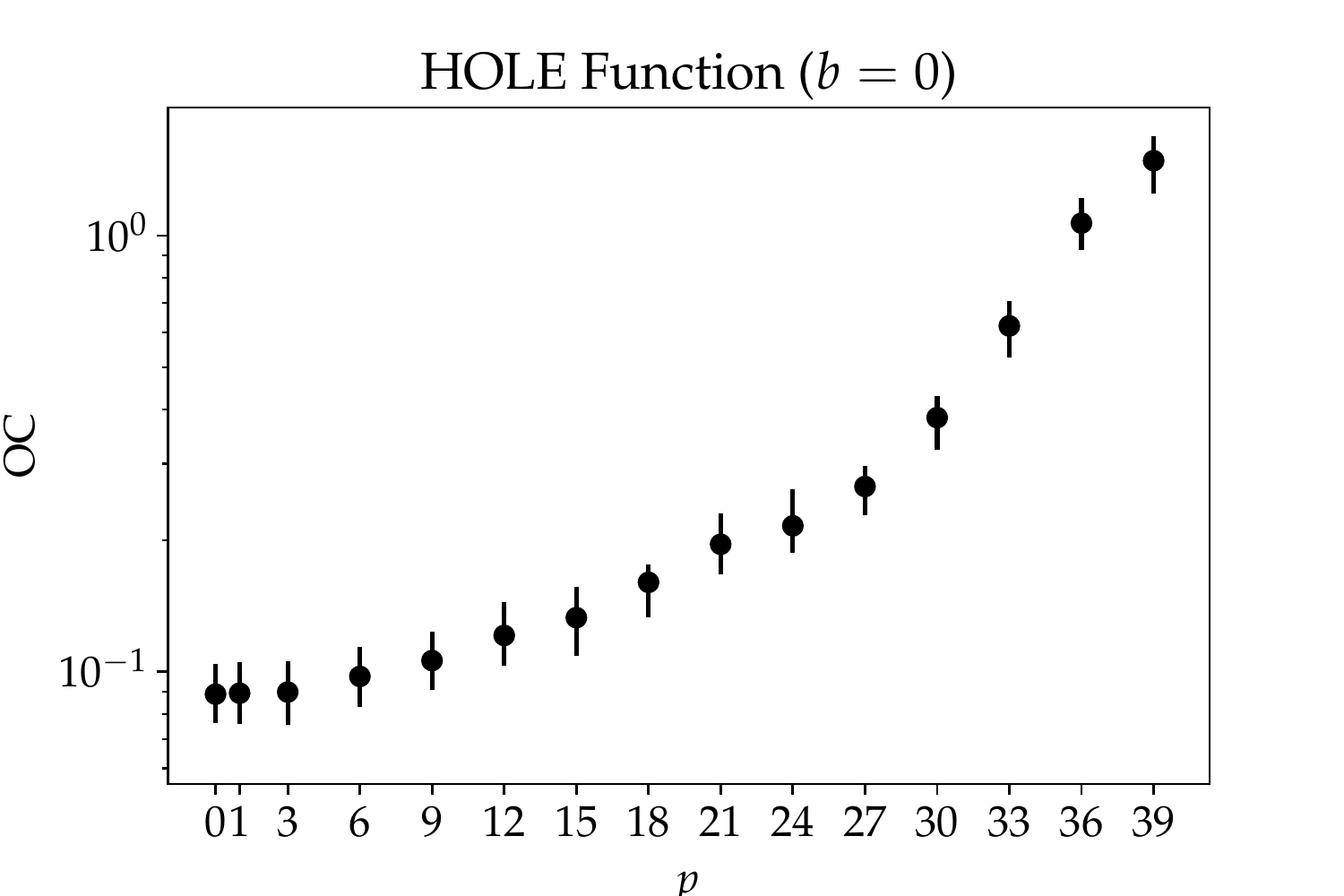}\\
		(b)\\
		\includegraphics[height=3.8cm]{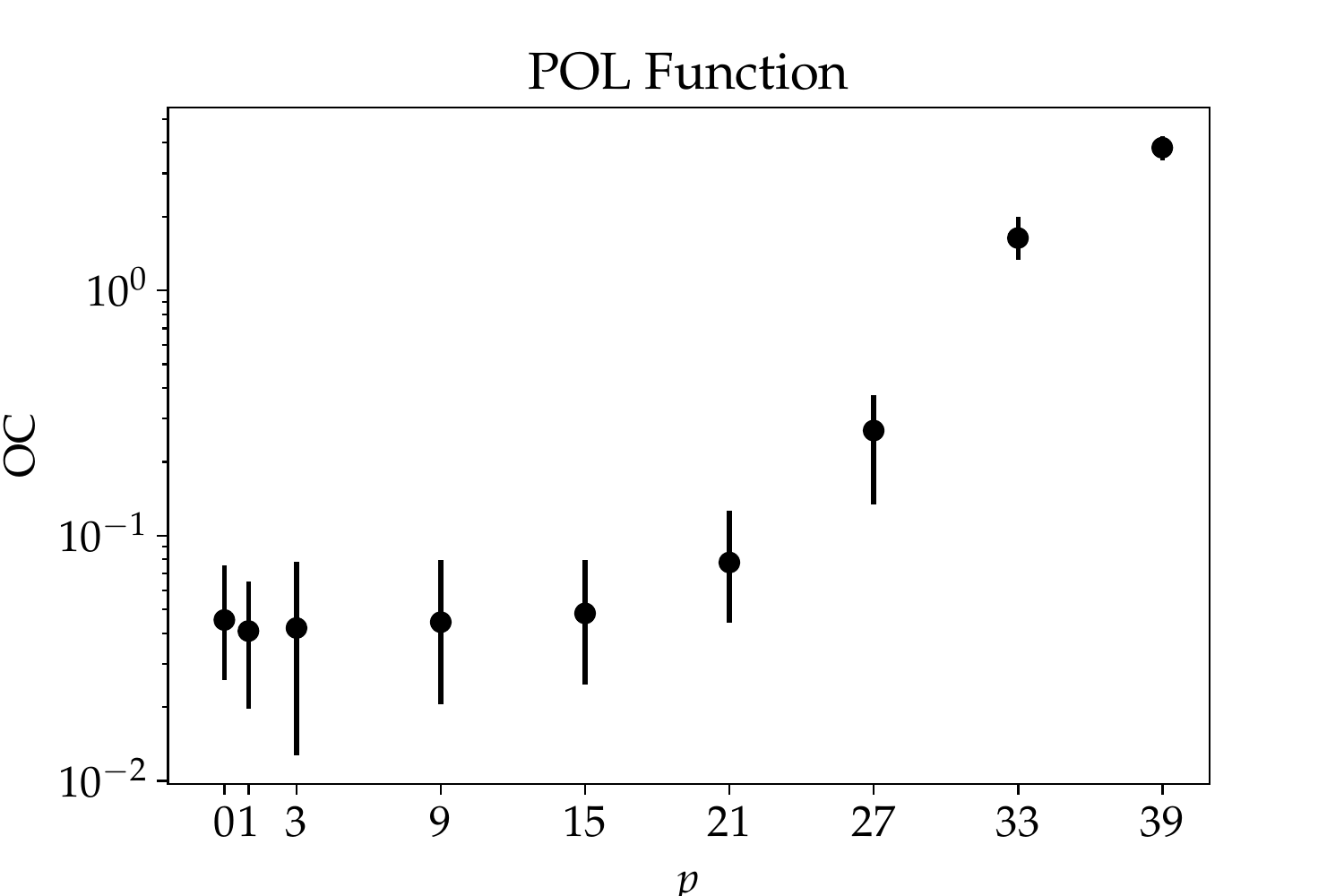}\\
		(c)\\
	\end{tabular}
    	\captionof{figure}{Final true utility of the generated solution set after $B=40$ iterations, depending on how many iterations before the end the DM was presented with an approximation of the front ($p$). True DM utility was \emph{linear}. Mean and 95\% CI for the OC.  (a) HOLE-1 function with $b>0$ (b) HOLE-2 function with $b=0$, (c) POL function.}
	\label{fig:HOLE_Ushape_B40_PLot}
\end{figure}

On the other hand, a restricted budget $\B$ results in a reduced accuracy of the approximated Pareto front shown to the DM and, as a consequence, the estimated parameter $\hat{\btheta}$ might not be a good approximation for $\btheta$. This would cause the proposed approach to focus on a less desirable direction during the last $\ksteps$ optimization steps. Therefore, under a restricted budget, the proposed approach may not be capable of taking advantage of several $\ksteps$. Figure~\ref{fig:HOLE_Ushape_B40_PLot} shows that decreasing the budget to $B=40$ results in better performance when the DM is shown the approximated front towards the end ($\ksteps=1$).

\section{Conclusion}
For the case of expensive multi-objective optimization, we show how the surrogate models generated by Bayesian optimization can be used not only to speed up optimization, but also to show an approximated continuous Pareto front to the DM once before the end of optimization. Then, the information on the most preferred solution can be used to focus the final iterations of the algorithm to try and find this predicted most preferred solution, or even a solution that dominates this solution. We argue that the additional cognitive effort for the DM should be small, but the benefit in terms of true utility to the DM is significant. We empirically demonstrate the effectiveness of the proposed approach on several test problems.

Future directions of research may include to explore this idea with other BO algorithms, and also to turn this into a fully interactive approach with multiple interactions with the DM during the optimization.


\section*{Acknowledgements}

The first author would like to acknowledge funding from ESTECO SpA and EPSRC through grant EP/L015374/1.
 	
 	 \bibliographystyle{ACM-Reference-Format}
  
    \bibliography{bibliography.bib}
    
    \appendix
\section{Appendix}\label{Appendix}

\subsection{HOLE Test function}

This problem is defined over $X = [-1,1]^{2}$ and has two objectives,

 \begin{center}
 \begin{tabular}{c } 
 \hline
 \textbf{Hole Problem}  \\ [0.5ex] 
 \hline\hline
 Design variables \\ 
 $x_{1} = [-1,1], \quad x_{2} = [-1,1]$\\
 \hline
 Function Parameters \\
 $q=0.2$ \quad$p =2$\quad $ d_{0} = 0.02$\\
 \hline
 Problem Hardness \\
 $h=0.2$\\
 \hline
 Translation\\
 $\delta = 1 - \sqrt{2}/2$\\
 $x_{1}' = x_{1} + \delta $\\
  $x_{2}' = x_{2} - \delta $\\
 \hline
 Rotation of 45 deg\\
 $\alpha = \pi / 4$\\
 $x_{1}'' = x_{1}'\cos(\alpha) + x_{2}'\sin(\alpha) $\\
 $x_{2}'' = -x_{1}'\sin(\alpha) + x_{2}'\cos(\alpha)  $\\
 \hline
 Scale of $\pi$\\
 $x_{1}''' = x_{1}'' \pi$\\
 $x_{2}''' = x_{2}'' \pi$\\
 \hline
 Change into problem Coordinates\\
 $u = \sin(x_{1}'''/2), \; u \in [-1,1]$\\
 $v = \sin^{2}(x_{2}'''/2), \; v \in [0,1]$\\
 \hline
 Apply Hardness\\
 $u' = u^{h} \quad\textrm{if}\quad  u\geq 0; \quad u'=-(-u)^{h} \quad\text{if}\quad u< 0$\\
 $v' = {v^{1/h}}$\\
 \hline
 Change into Problem Parameters\\
 $t = u', \quad t\in [-1,1]$\\
 $a = v'2p, \quad a \in [0, 2p]$\\
 \hline
 Other Parameter\\
 $b = (p-a)\exp(q) \quad\text{if}\quad a \leq p; b = 0 \quad\text{if}\quad a > p$\\
 $d = 1/2a + d_{0}$\\
 $c = q/d^{2}$\\
 \hline
 Objective Functions\\
 $\text{min }f_{1} = (t + 1)^{2} + a + b \exp[-(c-t)^{2}]$\\
$\text{min }f_{2} = (t - 1)^{2} + a + b \exp[-(c+t)^{2}]$\\
[1ex] 
 \hline
\end{tabular}
\end{center}

\subsection{POL Test function}
The POL problem has two input variables $x_{1}, x_{2} \in [-\pi, \pi]$ and two
objective functions:

\begin{align*}
\begin{split}
    \text{min }f_{1} &= 1+(a-b)^{2} + (c-d)^{2} 
\end{split}\\
\begin{split}
    \text{min }f_{2} &= (x_{1}+3)^{2} + (x_{2}+1) ^{2}
\end{split}
\end{align*}

with 

\begin{align*}
\begin{split}
    a &= 0.5 \sin(1) - 2.0 \cos(1) + 1.0 \sin(2) - 1.5 \cos(2)
\end{split}\\
\begin{split}
    b &= 0.5 \sin(x_{1}) - 2.0 \cos(x_{1}) + 1.0 \sin(x_{2}) - 1.5 \cos(x_{2})
\end{split}\\
\begin{split}
    c &= 1.5 \sin(1) - 1.0 \cos(1) + 2.0 \sin(2) - 0.5 \cos(2)
\end{split}\\
\begin{split}
    d &= 1.5 \sin(x_{1}) - 1.0 \cos(x_{1}) + 2.0 \sin(x_{2}) - 0.5 \cos(x_{2})
\end{split}
\end{align*}
\end{document}